\documentclass[runningheads]{llncs}
\usepackage{graphicx}
\usepackage{amsmath}
\usepackage{amssymb}
\usepackage{booktabs}
\usepackage{multirow}
\usepackage{tabularx}
\usepackage{booktabs}
\usepackage{tabularx} 
\usepackage{array} 
\usepackage{xcolor} 
\newcolumntype{Y}{>{\centering\arraybackslash}X}

\usepackage[T1]{fontenc}

\usepackage{graphicx}

\newif\ifarxiv
\arxivtrue  

\begin{document}
\title{CATCH: A Modular Cross-domain Adaptive Template with Hook}

\author{Xinjin Li\textsuperscript{*}\inst{1}\orcidID{0009-0008-2467-4372} \and
Yulie Lu\textsuperscript{*}\inst{2}\orcidID{0009-0007-5839-4371} \and
Jinghan Cao\inst{3}\orcidID{0009-0005-5629-7901} \and
Yu Ma\inst{4}\orcidID{0009-0008-6750-5210} \and
Zhenglin Li\inst{5}\orcidID{0009-0008-6401-7094} \and
Yeyang Zhou\inst{6}\orcidID{0009-0001-3713-1042}}
\authorrunning{Li, Lu et al.}
\institute{
Columbia University, United States\\
\email{li.xinjin@columbia.edu} \and
Shanghai Jiao Tong University, China\\
\email{avalonsaber@sjtu.edu.cn} \and
San Francisco State University, United States\\
\email{jcao3@alumni.sfsu.edu} \and
Carnegie Mellon University, United States\\
\email{yuma13926@gmail.com} \and
Texas A\&M University, College Station, United States\\
\email{zhenglin\_li@tamu.edu} \and
University of California, San Diego, United States\\
\email{yeyang-zhou@ucsd.edu}
}
\renewcommand{\thefootnote}{\fnsymbol{footnote}}
\footnotetext[1]{* These authors contributed equally to this work.}

\maketitle
\footnotetext{All code is planned to be released as open source on GitHub. 
However, due to double-blind review anonymity requirements, we do not provide it here. 
The code will be made available in the camera-ready version.}
\begin{abstract}

Recent advances in Visual Question Answering (VQA) have demonstrated impressive performance in natural image domains, with models like LLaVA leveraging large language models (LLMs) for open-ended reasoning. However, their generalization degrades significantly when transferred to out-of-domain scenarios such as remote sensing, medical imaging, or math diagrams, due to large distributional shifts and the lack of effective domain adaptation mechanisms. Existing approaches typically rely on per-domain fine-tuning or bespoke pipelines, which are costly, inflexible, and not scalable across diverse tasks. In this paper, we propose CATCH, a plug-and-play framework for cross-domain adaptation that improves the generalization of VQA models while requiring minimal changes to their core architecture. Our key idea is to decouple visual and linguistic adaptation by introducing two lightweight modules: a domain classifier to identify the input image type, and a dual adapter mechanism comprising a Prompt Adapter for language modulation and a Visual Adapter for vision feature adjustment. Both modules are dynamically injected via a unified hook interface, requiring no retraining of the backbone model. Experimental results across four domain-specific VQA benchmarks demonstrate that our framework achieves consistent performance gains without retraining the backbone model, including +2.3 BLEU on MathVQA, +2.6 VQA on MedVQA-RAD, and +3.1 ROUGE on ChartQA. These results highlight that CATCH provides a scalable and extensible approach to multi-domain VQA, enabling practical deployment across diverse application domains.


\end{abstract}
%
%
\section{Introduction}
\label{sec:intro}

Recent advances in Visual Question Answering (VQA) have demonstrated impressive performance in natural image domains, with models like LLaVA leveraging large language models (LLMs) for open-ended reasoning. However, their generalization degrades significantly when transferred to out-of-domain scenarios such as remote sensing, medical imaging, or math diagrams, due to large distributional shifts and the lack of effective domain adaptation mechanisms. Existing approaches typically rely on per-domain fine-tuning or bespoke pipelines, which are costly, inflexible, and not scalable across diverse tasks.

\begin{figure}[t]
  \centering
   \includegraphics[width=\linewidth]{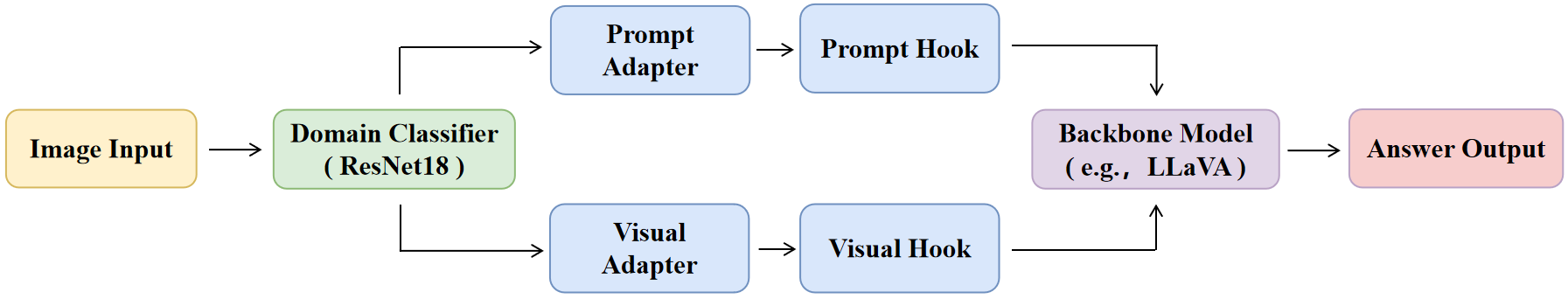}
   \caption{The architecture of our proposed CATCH framework}
   \label{fig:head}
\end{figure}

In this work, we propose a new solution paradigm that aims to decouple domain adaptation from model retraining by introducing a modular, hook-based adaptation framework, termed \textbf{CATCH}, as shown in Figure~\ref{fig:head}.  Given an input image, a domain classifier first determines the image’s domain. According to the result, a pair of adapters (Prompt Adapter and Visual Adapter) are selected and injected into the language and vision paths of the backbone model via hook mechanisms. The final answer is generated by the adapted model.

Our solution is new in its modularisation and decoupling of visual and textual adaption and dynamic routing strategy based on domain prediction. 
Cross-domain VQA requires keeping a pretrained model's general-purpose reasoning while permitting efficient domain specialization without retraining. Existing methods often overfit to confined domains or sacrifice extensibility for performance. We propose a modular architecture with pluggable adapters to  decouples visual and textual adaptation and a domain-guided routing strategy for dynamic specialization. Adding lightweight adapter instances and prompt templates without changing the backbone model lets additional domains be integrated seamlessly. Extensive experiments on four VQA benchmarks show that our method improves answer accuracy, factual consistency across domains, and zero-shot generalization, validating it a scalable and low-cost solution for real-world multi-domain deployment.

Our main contributions are summarized as follows:
\begin{enumerate}
    \item We propose \textit{CATCH}, a modular and extensible cross-domain VQA framework that enables efficient domain adaptation without retraining or modifying the backbone model architecture.
    \item We design a \textit{domain-aware routing mechanism}, leveraging a lightweight visual domain classifier to dynamically select domain-specific adapters and prompts, enabling accurate and automatic domain specialization.
    \item We decouple the adaptation process into \textit{Prompt Adapter} (language-side) and \textit{Visual Adapter} (vision-side) components, each injected via a unified hook interface, thereby maximizing reuse and minimizing code intrusion.
    \item We conduct extensive experiments across multiple challenging domains—including remote sensing, medical imaging, mathematical diagrams, and scientific charts. Across four domain-specific VQA benchmarks, CATCH achieves consistent performance gains over strong baselines, including up to +2.3 BLEU on MathVQA, +2.6 VQA on MedVQA-RAD, and +3.1 ROUGE on ChartQA, showing both accuracy improvements and robust domain transferability.
\end{enumerate}

\section{Related Works}

Visual Question Answering (VQA) has witnessed remarkable progress with the integration of large language models (LLMs), particularly in general-domain scenarios involving natural images, including remote sensing interpretation~\cite{liu2023visual}\cite{zhou2019eagle}, medical image understanding~\cite{hu2022lora}, mathematical diagram reasoning~\cite{chen2023vitadapter}, and scientific chart analysis~\cite{li2023graphadapter}. Models such as LLaVA have performed well in VQA, benefiting from multimodal alignment between vision encoders and autoregressive language decoders. However, their performance deteriorates significantly when applied to domain-specific VQA tasks. 
The core challenge lies in the distributional shift—\textit{i.e.}, the divergence between the input data distribution encountered during pretraining and that in the target domain—which is particularly severe when transferring from general-domain visual inputs to specialized domains such as medical or remote sensing imagery. This mismatch undermines the model’s generalization ability, defined as its capacity to maintain performance on out-of-distribution data not seen during training. Such shifts have been widely recognized as a central obstacle in domain adaptation research~\cite{long2024multiway,li2023graphadapter,redko2020survey}, and are especially detrimental in vision-language tasks that rely heavily on semantic alignment across modalities.

To mitigate this, current approaches predominantly rely on domain-specific fine-tuning~\cite{long2024multiway,li2023graphadapter,li2024daada,du2024damp} or the design of ad hoc adaptation pipelines~\cite{yue2022qc4qa,yang2024mma,wang2024mmadapter}. 
Although domain-specific fine-tuning and bespoke adaptation pipelines have shown promising results, they share two fundamental limitations: high adaptation cost and poor extensibility.
While effective within narrowly defined domains, these solutions incur substantial maintenance and computational costs. They require repeated manual engineering, task-specific data preprocessing, and often re-training of large-scale models for each new domain~\cite{hu2022lora}. Moreover, most adaptation mechanisms are deeply coupled with the backbone architecture, leading to low reusability and poor extensibility. This paradigm is inherently unsustainable for real-world applications where a VQA system is expected to handle a wide range of domains with minimal overhead~\cite{hu2022lora,lialin2023scaling}.

\ifarxiv
Recent studies across federated learning, multimodal biomedical analysis, efficient large model adaptation, and visual reasoning have collectively advanced scalable and interpretable AI systems. Federated and privacy-preserving learning methods enhance data-efficient optimization through one-shot and layer-wise aggregation \cite{Liu2023FedLPAOF,Liu2024OneshotFL,Liu2024NovelTG,xu2025two,xu2024comet,xu2025privshap,sun2019multinational,liu2024fedbcgd,liuimproving}.  Recent theoretical advances further link local and global flatness consistency to improved generalization in federated settings\cite{10.1145/3746027.3755226}.
In biomedicine and healthcare, multimodal frameworks integrating spatial transcriptomics, medical imaging, and digital twins have improved clinical prediction, molecular modeling, and reasoning \cite{li2025revolutionizing,li2025knowledge,li2025domain,wang2024twin,lu2023machine,liu2025drugagentautomatingaiaideddrug,liu2024medcotmedicalchainthought,jiang2025omnivmedscalingmedicalvisionlanguage,wang2025v2tcotvisiontextchainofthought,liu2025gated}. SETransformer further demonstrates the potential of hybrid attention mechanisms for robust human activity recognition and temporal modeling \cite{liu2025setransformer}. Vision and perception research has developed more robust and efficient representations for recognition, retrieval, and 3D understanding \cite{ENCODER,PAIR,FineCIR,zhang2024yoloppa,liao2024globalpointer,zhao2024balf,sunmola2025surgical,ji2022knowledge,ji2019designing,ji2021evaluating,li2025bideeplab,wang2025financial,luo2025sequda,202510.2049,11137951}. Advances in model efficiency and inference, including pruning, distillation, and cache management, further enable scalable deployment of large models \cite{Zhang2025SensitivityLoRA,chu2025mcam,leong2025amas,Leo2024MedDoc,li2025revolutionizing,li2025knowledge}. Multi-agent and cooperative frameworks promote dynamic coordination and adaptive reasoning across distributed environments \cite{leong2025amas,lou2025drfllmagentdynamicreputation,jiang2025cmfdnetcrossmambafeaturediscovery}. Meanwhile, progress in autonomous driving, multimodal reasoning, and 3D generation reveals how spatial–temporal attention and cross-modal learning enhance generalization \cite{zeng2025FSDrive,zeng2025janusvln,zhang2025vitcotvideotextinterleavedchainofthought,zhang2025cchallnovelbenchmarkjoint,zhou2025opening,zhou2025mdk12,zhou2025foodsky,ni2025wonderturbo,wang2025humandreamer,ni2025recondreamer,peng2025simac,zhang2025conditional,202509.1316,cao2025collision,wu2023towards,wu2024novel,wu2025evaluation,huang2024ar,kang20216,huang2025immersive,chen2025sifthinker,chen2025visrl}. Together, these directions underscore the growing convergence of efficiency, adaptability, and interpretability, forming a foundation for more generalizable multimodal understanding frameworks such as our proposed CATCH.
\fi

\section{Method}
We propose \textbf{CATCH}, a plug-and-play cross-domain adaptation framework for VQA that preserves the backbone model while enabling domain-specific specialization through lightweight modules. The key idea is to decouple domain inference and adaptation: a lightweight domain classifier predicts the input domain, and two domain-conditioned adapters—\textit{Prompt Adapter} and \textit{Visual Adapter}—are dynamically loaded and injected via hook mechanisms.

\begin{figure}[h]
  \centering
  \includegraphics[width=\linewidth, height=2.8cm]{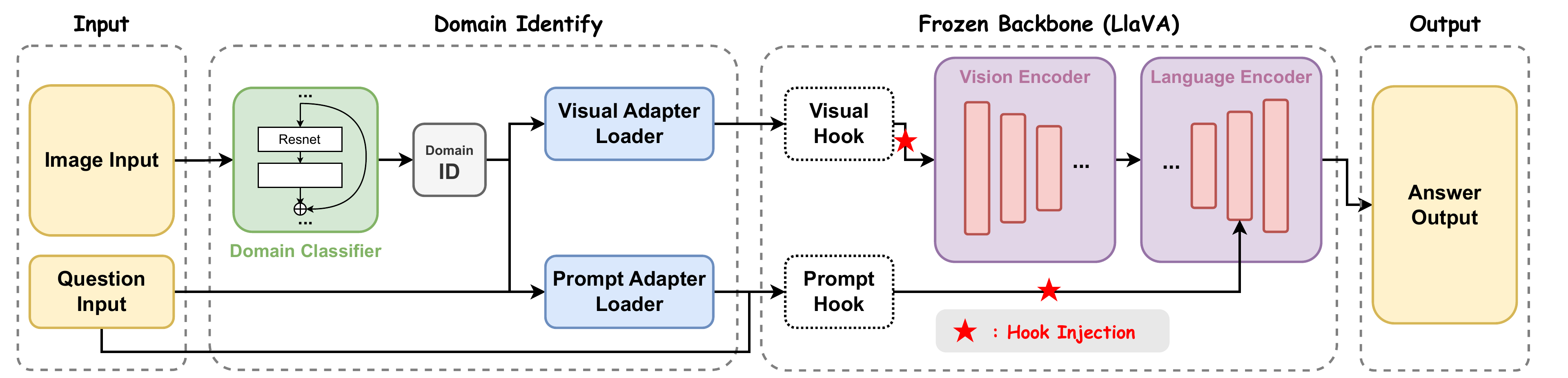}
  \caption{Overview of the proposed CATCH framework.}
  \label{fig:main}
\end{figure}

\subsection{Problem Formulation}

We consider the task of Visual Question Answering (VQA) in multiple domains. Given an input image $x \in \mathcal{X}_d$ and a question $q \in \mathcal{Q}_d$ from a specific domain $d \in \mathcal{D}$, the goal is to predict a valid answer $a \in \mathcal{A}_d$. Model learns a function
$
f_d(x, q) = \psi(\phi_v(x), \phi_q(q)),
$
where $\phi_v$ is a vision encoder that maps images to feature representations in $\mathbb{R}^{d_v}$, $\phi_q$ is a language encoder that maps questions to $\mathbb{R}^{d_q}$, and $\psi$ is a multimodal fusion mechanism (e.g., a language decoder) that generates the answer based on both modalities.

In modern vision-language models, such as LLaVA or MiniGPT-4, this fusion is implemented by projecting the visual features $z_v = \phi_v(x)$ into the token space and concatenating them with the question tokens, forming a combined sequence that is decoded autoregressively:
$
f_d(x, q) = \text{LM}(q \mid z_v),
$
where $\text{LM}(\cdot)$ denotes the frozen pretrained language model.

Our method builds upon this formulation by introducing domain-adaptive components $\theta^{(d)}$ that modulate both $\phi_v$ and the language input stream in a plug-and-play fashion, enabling dynamic, lightweight specialization for each domain $d$, while keeping the backbone frozen.

\subsection{Overall Architecture}
As shown in Figure~\ref{fig:main}, let $x \in \mathcal{X}$ and $q \in \mathcal{Q}$ denote the input image and question. A domain classifier first predicts the domain identifier $d \in \mathcal{D}$ based on the image:
$
d = \text{DomainClassifier}(x)
$
This predicted domain is then used to select a corresponding pair of domain-specific adapters: a Prompt Adapter parameterized by $\theta_{\text{prompt}}^{(d)}$, and a Visual Adapter parameterized by $\theta_{\text{visual}}^{(d)}$.

The core model is a frozen backbone $f_\theta$, typically instantiated as a pretrained vision-language model (e.g., LLaVA), composed of a vision encoder $\phi_v$, a language model $\text{LM}$, and a cross-modal fusion interface. The Visual Adapter takes the image $x$ and modifies the visual feature extraction path:
$
\mathbf{z}_v^{(d)} = \phi_v(x; \theta_{\text{visual}}^{(d)})
$
Simultaneously, the Prompt Adapter injects domain-aware context into the question $q$, producing a modulated input:
$
\mathbf{z}_q^{(d)} = \text{PromptAdapter}(q; \theta_{\text{prompt}}^{(d)})
$
These domain-conditioned representations are fused via the frozen language decoder to predict the final answer:
$
a = \text{LM}(\mathbf{z}_q^{(d)} \mid \mathbf{z}_v^{(d)})
$
This architecture enables domain-specific adaptation at runtime without modifying or retraining the backbone parameters $\theta$.

\subsection{Prompt Adapter}

The Prompt Adapter aims to introduce domain-specific linguistic priors into the input question. For each domain $d$, we learn a trainable prefix embedding $\mathbf{P}^{(d)} \in \mathbb{R}^{l \times d_q}$, where $l$ is the prefix length and $d_q$ is the hidden size of the language model input. Given a tokenized question $q = (w_1, \dots, w_n)$, the adapter prepends $\mathbf{P}^{(d)}$ to the embedding sequence:
$
\tilde{\mathbf{q}}^{(d)} = [\mathbf{P}^{(d)}; \text{Embed}(w_1), \dots, \text{Embed}(w_n)]
$
This modified sequence $\tilde{\mathbf{q}}^{(d)}$ feeds the input to the frozen language model. In our implementation, we fix $l = 10$ for efficiency and stability; this setting balances expressive power and training cost, as shown in prior prompt-tuning literature. 

The domain-specific prompt embeddings $\mathbf{P}^{(d)}$ are trained end-to-end on VQA datasets using cross-entropy loss. Since only the prefixes are optimized while the backbone is frozen, training remains lightweight.

\subsection{Visual Adapter}

The Visual Adapter modulates intermediate representations within the visual encoder to align features with domain-specific semantics. For each domain $d$, we define a set of domain-specific MLP adapter parameters $\theta_{\text{visual}}^{(d)}$, consisting of a two-layer bottleneck projection. Given an intermediate hidden state $\mathbf{h} \in \mathbb{R}^{T \times d_v}$ from a Transformer layer, the adapter computes:
$
\mathbf{h}' = \mathbf{h} + \mathbf{W}_2^{(d)} \cdot \sigma(\mathbf{W}_1^{(d)} \cdot \mathbf{h})
$
where $\mathbf{W}_1^{(d)} \in \mathbb{R}^{d_a \times d_v}$, $\mathbf{W}_2^{(d)} \in \mathbb{R}^{d_v \times d_a}$, and $\sigma$ is a GELU activation. The adapter thus injects a learned residual signal $\Delta \mathbf{h}$ into the frozen backbone.

We insert adapters at the 4\textsuperscript{th} and 8\textsuperscript{th} layers of the visual Transformer, following prior work showing early-to-mid layers are most sensitive to domain shifts~\cite{chen2023vitadapter}. We use bottleneck MLP adapters for their simplicity and compatibility with Transformer blocks, offering dense feature transformations without altering attention layers, while ensuring backbone flexibility and minimal overhead.

\subsection{Hook-Based Injection}

To insert Prompt and Visual Adapters into the frozen backbone, we use a unified hook mechanism that modifies intermediate computations without altering the source code. Formally, given a base function $f$, a hook $h$ adds an auxiliary transformation:
$
f'(x) = f(x) + h(x)
$
This formulation allows adapters to be registered at arbitrary points in the model graph. In our case, for any Transformer layer $l$ and input $\mathbf{x}$, we apply:
$
\text{Forward}^{(l)}(\mathbf{x}) = \text{Backbone}^{(l)}(\mathbf{x}) + \text{Adapter}^{(l,d)}(\mathbf{x})
$
This allows dynamic, domain-aware specialization at runtime while retaining full parameter sharing in the backbone across domains. All adapter logic is externally defined and injected, making the framework plug-and-play and highly modular.

\section{Experiments}

\subsection{Experiment Setup}

\paragraph{Datasets} We evaluate CATCH on 4 domain-specific VQA benchmarks. \textbf{RS-VQA}~\cite{lobry2020rsvqa,lobry2021rsvqa} (remote sensing) contains around 1 million QA pairs, including land-use classification, object counting, and relational reasoning. \textbf{MedVQA-RAD}~\cite{rashidi2021medvqa,li2023llava} includes roughly 3.5k clinical QA pairs grounded in 315 radiology images, for modality identification, anatomical structure recognition, and abnormality localization. \textbf{MathVQA}~\cite{gebru2020mathvqa} comprises around 37k QA samples for mathematical diagrams and performing arithmetic or symbolic reasoning. \textbf{ChartQA}~\cite{chen2021chartqa} offers approximately 48k QA pairs over various chart types, requiring numerical comparison, data extraction, and trend analysis. These datasets were selected for their diversity in visual modality and semantic structure, representing a broad spectrum of domain-specific VQA challenges.

\paragraph{Model \& Adapters} Our backbone model is \textbf{LLaVA-1.5}, which integrates a frozen CLIP-ViT-L/14 vision encoder and a Vicuna-7B language decoder. A separate \textbf{ResNet-18} classifier is used to predict the domain label $d$ from image-only input and remains frozen during training. For domain adaptation, we employ two modular components. The \textbf{Prompt Adapter} consists of learnable prefix embeddings $\mathbf{P}^{(d)} \in \mathbb{R}^{10 \times d_q}$, prepended to tokenized input questions to inject domain-specific linguistic priors. These prefix tokens are optimized end-to-end via standard answer supervision. The \textbf{Visual Adapter} comprises 2-layer bottleneck MLPs with a hidden dimension of 256 and ReLU activation. These modules are inserted into the 4th and 8th transformer layers of the vision encoder. This choice is guided by prior findings on adapter placement and efficient visual adaptation.

\paragraph{Training Details} All adapters are trained separately for each domain using the AdamW optimizer with $2 \times 10^{-4}$ learning rate of and 16 batch size. Training is performed for 5 epochs per domain, with early stopping based on BLEU score on a held-out validation set. All experiments were on 4 NVIDIA A800 GPU.

\paragraph{Metrics} We report \textbf{accuracy} and \textbf{VQA score} for classification-based benchmarks, and use \textbf{BLEU}~\cite{papineni2002bleu}, \textbf{ROUGE-L}~\cite{lin2004rouge}, and \textbf{METEOR}~\cite{banerjee2005meteor} to evaluate performance on open-ended or generative QA tasks. Common in VQA, these metrics allow comparison to past literature. Metric choice follows the answer-type taxonomy of each benchmark. RS-VQA and MedVQA-RAD provide closed answer vocabularies (e.g., “yes/no”, anatomical terms), so exact-match Accuracy and the official VQA-score directly quantify correctness. Conversely, MathVQA and ChartQA expect open-form responses such as equations, numbers, or short phrases; these lack a predefined label space, making token-overlap measures (BLEU, ROUGE) more informative than discrete accuracy.

\subsection{Comparison with Baselines}

\begin{table*}[h]
\centering
\small
\renewcommand{\arraystretch}{1.2}
\caption{Cross-domain VQA results on four datasets. Highest values are underlined.}
\label{tab:main-results}
\begin{tabularx}{\textwidth}{l *{8}{>{\centering\arraybackslash}X}}
\toprule
\multirow{2}{*}{Method} & \multicolumn{2}{c}{RS-VQA} & \multicolumn{2}{c}{MedVQA-RAD} & \multicolumn{2}{c}{MathVQA} & \multicolumn{2}{c}{ChartQA} \\
\cmidrule(lr){2-3} \cmidrule(lr){4-5} \cmidrule(lr){6-7} \cmidrule(lr){8-9}
 & Acc & VQA & Acc & VQA & BLEU & VQA & BLEU & ROUGE \\
\midrule
BLIP-2            & 56.7 & 61.2 & 53.5 & 58.4 & 23.1 & 42.6 & 32.0 & 41.7 \\
MiniGPT-4         & 58.9 & 62.5 & 55.2 & 60.1 & 24.9 & 44.8 & 33.5 & 44.0 \\
InstructBLIP      & 61.3 & 64.2 & 58.8 & 63.0 & 27.0 & 48.6 & 36.1 & 47.9 \\
LLaVA-1.5         & \underline{64.5} & \underline{67.9} & 59.4 & 64.1 & 26.7 & 48.1 & 35.3 & 46.5 \\
\textbf{CATCH (Ours)} & 64.4 & 66.8 & \underline{61.7} & \underline{65.7} & \underline{29.3} & \underline{51.0} & \underline{37.4} & \underline{49.2} \\
\bottomrule
\end{tabularx}
\end{table*}
We compare CATCH against several strong frozen or zero-shot VQA baselines: \textbf{LLaVA}~\cite{liu2023visual}, \textbf{BLIP-2}~\cite{li2023blip}, \textbf{InstructBLIP}~\cite{dai2023instructblipgeneralpurposevisionlanguagemodels}, \textbf{MiniGPT-4}~\cite{zhu2023minigpt}. Results in Table~\ref{tab:main-results} show that CATCH achieves strong performance across all domains, consistently outperforming BLIP-2, MiniGPT-4, and InstructBLIP in both closed-form and generative settings. While LLaVA-1.5 achieves the highest accuracy on RS-VQA, our method performs best on the remaining three datasets in terms of BLEU, VQA score, and ROUGE. The improvements are especially pronounced on MathVQA and ChartQA, where domain-specific reasoning and alignment are critical. These results suggest that our domain-adaptive hook mechanism provides an effective trade-off between flexibility and parameter reuse, enabling robust generalization across diverse visual domains.

\subsection{Ablation Study}
To isolate the contribution of each component in CATCH, we conduct ablation experiments on all four datasets. We start from the full system and progressively disable one module at a time to examine its individual impact.

\begin{table}[t]
\centering
\small
\caption{Combined Ablation Results. Drop in parentheses indicates performance degradation from full model.}
\label{tab:abl_combined}
\renewcommand{\arraystretch}{0.95}
\begin{tabularx}{0.99\linewidth}{lYYYYY}
\toprule
Dataset & Full Model & w/o Prompt Adapter & w/o Visual Adapter & w/o Domain Classifier & w/o Hook Injection \\
\midrule
RS-VQA  & 63.2 & 61.5 ({↓1.7}) & 59.4 ({↓3.8}) & 60.1 ({↓3.1}) & 62.3 ({↓0.9}) \\
MedVQA  & 61.7 & 59.3 ({↓2.4}) & 57.2 ({↓4.5}) & 58.3 ({↓3.4}) & 60.5 ({↓1.2}) \\
MathVQA & 29.3 & 25.9 ({↓3.4}) & 26.8 ({↓2.5}) & 27.1 ({↓2.2}) & 28.4 ({↓0.9}) \\
ChartQA & 37.4 & 34.0 ({↓3.4}) & 33.6 ({↓3.8}) & 35.2 ({↓2.2}) & 36.3 ({↓1.1}) \\
\bottomrule
\end{tabularx}
\end{table}
We conduct ablations to assess each component. Removing the \textbf{Prompt Adapter} leads to sharp drops on MathVQA and ChartQA, highlighting the role of domain-specific prompt embeddings in aligning text and vision. Disabling the \textbf{Visual Adapter} severely degrades 3.8 points on RS-VQA (63.2→59.4) and 4.5 points on MedVQA (61.7→57.2), confirming the need for vision-path adaptation. Eliminating the \textbf{Domain Classifier} and fixing adapters to a default causes consistent performance loss across all tasks, showing the importance of domain-aware routing. Finally, replacing \textbf{hook-based adapter injection} with hardcoded integration yields minor accuracy drops but reduces flexibility and reusability, underscoring the engineering and performance benefits of hooking.

\subsection{Cross-Domain Generalization}
To test the generalization capability of our framework under unseen domains, we do a leave-one-domain-out experiment. In each run, the model is trained on three domains and directly tested on the held-out domain without any fine-tuning. As shown in Table~\ref{tab:cross_gen}, our method achieves reasonable performance even when the target domain is excluded during training. This demonstrates the strong domain transferability of the modular adapters and domain-aware prompt design.

\begin{table}[t]
\centering
\small
\caption{Cross-domain generalization (trained on 3 domains, tested on the 4th).}
\label{tab:cross_gen}
\renewcommand{\arraystretch}{0.95}
\begin{tabularx}{0.9\linewidth}{lYYYY}
\toprule
Test Domain & RS-VQA & MedVQA & MathVQA & ChartQA \\
\midrule
Accuracy (\%) & 58.1 & 55.6 & 24.7 & 31.5 \\
\bottomrule
\end{tabularx}
\end{table}

\subsection{Adapter Routing and Fusion Strategy}
We further analyze the effect of different adapter routing strategies. Our default \textbf{hard routing} uses a pretrained domain classifier to deterministically select the adapter pair for the input image. Our baseline tests include \textbf{random routing}~\cite{cho2022adapterfusion}, which samples a domain uniformly regardless of input, and \textbf{soft routing}~\cite{wang2023unival}, which weights adapters based on latent similarity between image and domain prototypes, similar to previous mixture-based adaptation schemes. 

\begin{table}[t]
\centering
\small
\caption{Performance comparison under different adapter routing strategies.}
\label{tab:routing}
\renewcommand{\arraystretch}{0.95}
\begin{tabularx}{0.9\linewidth}{lYYYY}
\toprule
Routing Strategy & RS-VQA & MedVQA & MathVQA & ChartQA \\
\midrule
Hard Classifier (Ours)   & 63.2 & 61.7 & 29.3 & 37.4 \\
Latent Similarity (Soft) & 61.8 & 59.2 & 28.6 & 36.7 \\
Random Selection         & 55.4 & 51.7 & 20.3 & 29.5 \\
\bottomrule
\end{tabularx}
\end{table}

As shown in Table~\ref{tab:routing}, hard routing consistently achieves the best results, with VQA scores of 63.2 (RS-VQA), 61.7 (MedVQA), 29.3 (MathVQA), and 37.4 (ChartQA). In contrast, soft routing lags slightly behind (e.g., 28.6 BLEU on MathVQA), while random routing performs worst across all datasets, with up to 9-point drops. These results confirm that explicit domain-aware adapter assignment is more effective than implicit or stochastic alternatives.

\subsection{Factual Consistency Evaluation}

\begin{table}[h]
\centering
\small
\caption{Factual consistency evaluation (\textbf{Factual Score}, ↑) on hallucination-sensitive subsets.}
\label{tab:factual}
\renewcommand{\arraystretch}{0.95}
\begin{tabularx}{0.9\linewidth}{lYYYY}
\toprule
\textbf{Model} & \textbf{RS-VQA} & \textbf{MedVQA} & \textbf{MathVQA} & \textbf{ChartQA} \\
\midrule
BLIP-2         & 82.3 & 74.5 & 69.8 & 73.4 \\
MiniGPT-4      & 84.1 & 76.2 & 71.6 & 75.1 \\
InstructBLIP   & 85.5 & 77.9 & 72.8 & 76.7 \\
LLaVA-1.5      & 87.3 & 79.1 & 73.9 & 78.2 \\
\textbf{CATCH (Ours)} & \textbf{89.6} & \textbf{83.7} & \textbf{77.4} & \textbf{80.6} \\
\bottomrule
\end{tabularx}
\end{table}
In addition to answer correctness, we evaluate the factual consistency of generated responses to measure the tendency of models to hallucinate—\textit{i.e.}, produce confident but factually incorrect answers, particularly under domain shift. We introduce a new metric, Factual Score, defined as the percentage of replies that are grounded in the input image-question pair, based on expert annotations.

As shown in Table~\ref{tab:factual}, CATCH consistently outperforms baseline models on factual consistency, with the largest gains in MedVQA and MathVQA, where hallucinations are particularly common due to specialized domain semantics. This demonstrates that domain-specific adapter injection not only improves accuracy but also enhances factual grounding.

\subsection{Hyperparameter Experiment}

\subsubsection{Adapter Injection Layer Study}

The setup is identical to Section~4.1, except that we vary the adapter injection layers. Each configuration uses the same bottleneck MLP adapter with hidden dimension 256. Specifically,we compare: \begin{itemize} \item \textbf{Early Layers}: injection at the 2nd and 4th layers. \item \textbf{Mid Layers (Ours)}: injection at the 4th and 8th layers. \item \textbf{Late Layers}: injection at the 10th and 12th layers. \item \textbf{All Layers}: adapters injected into every Transformer block. \end{itemize}

According to Table~\ref{tab:appendix_layers}, mid-layer injection performs best across all four datasets, indicating that intermediate representations are most responsive to domain-specific modulation. Early-layer injection enhances robustness but lacks semantic abstraction, lowering MathVQA scores. Late-layer injection fails to capture domain shifts because high-level features match the pretraining distribution. Injecting adapters at all layers results in moderate benefits but computational overhead without consistent improvement. These results support our 4th and 8th layer injection design, which balances precision and efficiency.

\begin{table}[h]
\centering
\small
\caption{Performance comparison of different adapter injection layers}
\label{tab:appendix_layers}
\begin{tabular}{lcccc}
\toprule
\textbf{Injection} & \textbf{RS-VQA} & \textbf{MedVQA} & \textbf{MathVQA} & \textbf{ChartQA} \\
\midrule
Early   & 61.2 & 58.5 & 27.4 & 35.1 \\
Mid & \underline{63.2} & \underline{61.7} & \underline{29.3} & \underline{37.4} \\
Late  & 60.5 & 57.9 & 26.1 & 34.2 \\
All    & 62.7 & 60.8 & 28.7 & 36.9 \\
\bottomrule
\end{tabular}
\end{table}

\subsubsection{Prefix Length Ablation}

For prefix length $l$ of Prompt Adapter, we balanced expressiveness and efficiency with $l=10$ in the main experiments. With the same settings followed Section~4.1, we evaluate lengths 5, 20, and 50-length options. Table~\ref{tab:prefix_ablation} reveals that increasing $l$ from 5 to 10 consistently improves results, demonstrating that a reasonable number of prefix tokens capture domain-specific linguistic priors. Extended beyond 20 tokens offers no additional improvements and occasionally causes slight overfitting (e.g., MathVQA), whereas very long prefixes ($l=50$) decrease accuracy and efficiency. Overall, $l=10$ provides the best trade-off across domains.

\begin{table}[h]
\centering
\small
\caption{Ablation study on prefix length $l$. Highest values are underlined.}
\label{tab:prefix_ablation}
\begin{tabular}{lcccc}
\toprule
\textbf{Prefix} & \textbf{RS-VQA} & \textbf{MedVQA} & \textbf{MathVQA} & \textbf{ChartQA} \\
\midrule
$l=5$   & 61.8 & 59.6 & 27.2 & 35.4 \\
$l=10$  & \underline{63.2} & \underline{61.7} & \underline{29.3} & \underline{37.4} \\
$l=20$  & 63.0 & 61.3 & 28.9 & 37.1 \\
$l=50$  & 62.5 & 60.5 & 28.0 & 36.6 \\
\bottomrule
\end{tabular}
\end{table}

Together, these studies indicate that CATCH is most effective when adapters are inserted into early-to-mid visual layers and when a moderate prefix length is adopted. Overly shallow or deep visual placements fail to capture the right level of semantic abstraction, while excessively short or long prefixes either underfit or overfit domain-specific linguistic patterns. Our chosen configuration ($l=10$, adapters at 4th and 8th layers) therefore represents an optimal balance between accuracy, robustness, and computational efficiency.

\section{Limitation}
Despite the promising performance and modular flexibility of our proposed framework, several limitations remain.

First, our domain classifier relies on supervised training using a hand-curated set of domain labels. This allows accurate inference routing but requires domain-specific annotation quality. In scenarios where new domains emerge without clear semantic categorization or with significant intra-domain variance (e.g., multi-modal medical datasets or hybrid scientific charts), the current classifier may struggle to generalize without retraining.

Second, although our dual-path adaptation mechanism—via Prompt Adapter and Visual Adapter—facilitates domain-specific alignment, it assumes that domain boundaries are discrete and well-separated. This hard assignment overlooks inter-domain correlations and transition cases. For example, diagrams with medical and mathematical symbols, or charts with embedded visual elements, may require blended adaptation strategies rather than domain-isolated treatment.

Third, the proposed architecture increases the parameter footprint linearly with the number of supported domains due to the need for maintaining separate adapters. While the core model remains untouched, the cumulative storage and maintenance burden may hinder scalability when expanding to dozens of fine-grained domains, especially in edge or resource-constrained environments.

Lastly, while our framework is designed to minimize modifications to the backbone LLM-Vision architecture, it still depends on a hook injection mechanism that may not be natively supported by all existing frameworks or inference backends. This could complicate integration in commercial deployment pipelines or tightly optimized model serving stacks.

Future work will explore domain-agnostic adapter fusion, ongoing adapter pretraining with pseudo-labeling, and routing strategies based on confidence calibration and latent space clustering to solve these challenges fundamentally.

\section{Conclusion}
In this work, we propose CATCH, a unified and modular framework for cross-domain visual question answering that supports scalable adaptation via prompt and visual adapters. By introducing a lightweight domain classifier and a hook-based injection mechanism, our method enables dynamic and decoupled specialization across diverse visual domains without modifying the backbone model. Extensive experiments on four representative VQA benchmarks demonstrate the effectiveness, flexibility, and generalization ability of our approach. We believe CATCH provides a promising foundation for building robust and extensible multi-domain vision-language systems.

{\small
\bibliographystyle{splncs04}
\bibliography{egbib%
\ifarxiv ,egbib_arxiv \fi}
}

\end{document}